# mshw, a forecasting library to predict short-term electricity demand based on multiple seasonal Holt-Winters

Oscar Trull[a], J.Carlos García-Díaz[a], A. Peiró-Signes[b]

[a]*Department of Applied Statistics, Operational Research and Quality, Universitat Politècnica de València, E46022 Spain*

[b]*Management Department, Universitat Politècnica de València, E46022 Spain*

**Abstract**

Transmission system operators have a growing need for more accurate forecasting of electricity demand. Current electricity systems largely require demand forecasting so that the electricity market establishes electricity prices as well as the programming of production units. The companies that are part of the electrical system use exclusive software to obtain predictions, based on the use of time series and prediction tools, whether statistical or artificial intelligence. However, the most common form of prediction is based on hybrid models that use both technologies. In any case, it is software with a complicated structure, with a large number of associated variables and that requires a high computational load to make predictions. The predictions they can offer are not much better than those that simple models can offer. In this paper we present a MATLAB® toolbox created for the prediction of electrical demand. The toolbox implements multiple seasonal Holt–Winters exponential smoothing models and neural network models. The models used include the use of discrete interval mobile seasonalities (DIMS) to improve forecasting on special days. Additionally, the results of its application in various electrical systems in Europe are shown, where the results obtained can be seen. The use of this library opens a new avenue of research for the use of models with discrete and complex seasonalities in other fields of application.

## 1. Introduction

The evolution of national electrical systems in Western countries towards deregulation and a free market has forced all entities to improve their management. Currently, the markets are organized by producers, sellers and a transmission system operator (TSO) who is in charge of establishing the operating conditions, and with it, the cost of energy [1]. These TSOs may or may not be owned by governments, but they work independently in search of the best energy efficiency. They are in charge of matching supply and demand, and thus programming the production units. However, and prior to this programming, the producers make offers in the electricity market by offering quantities of energy at a price. In the same way, marketers offer energy purchases at a cost. The TSO is in charge of balancing these offers to establish which units will be the active ones to generate energy, and the price of energy is established. This responsibility has turned the TSOs as the main elements in the electricity system.

This business model moves in Spain around 40 billion euros, in Germany more than 70 billion and in France more than 80 billion and it depends enormously on these predictions [2–4]. The forecasts thus provided by TSOs must be accurate. Hong estimates that for a 1000MW production unit, an improve of 1% in the forecast error can reduce losses of up to $300,000 per year while analyzing short-term load forecasting (STLF) and 500,000$ when analyzing long-term load forecasting [5].

Due to the fact that predictions play a fundamental role in the electricity system, the entities that are part of the electricity systems have opted for the development of specific software for the prediction of demand and prices, each one for its field of action. There is a wide variety of software provided by specialized companies [6], although on many occasions, TSOs prefer to develop their own [7,8]. These models have a huge number of parameters and factors to calculate, in addition to having to rework the series to be able to carry out the calculations. The models offer very good results, although they require a huge investment and require highly capable equipment.

In this article we present a novelty MATLAB® toolbox specifically designed for electricity demand forecasting, in which its main objective has been the simplicity of the models, as well as robust results.

This toolbox includes multiple seasonal Holt-Winters (nHWT) models, prediction models based on component decomposition, and artificial neural network (ANN) models. In addition, the possibility of using discrete interval moving seasonalities (DIMS) to make predictions on special days is included.

The paper is organized as follows. Section 2 surveys the previous published works related to this paper. Section 3 shows the models used in this toolbox. Section 4 describes the architecture of the toolbox. Section 5 presents some cases published using this software. Finally, Section 6 summarizes the conclusions and the future research.

## 2. Related work

The generation of electricity for public use dates back to the end of the 19th century with the installation of the first plants in Godalming (UK) and Pearl Street (US). The first plants produced without the need to manage demand, due to the low installed capacity. But with the growth in the use of electricity, and with the installation of more and more production plants, it was necessary to focus in knowing how much energy should be produced. The growing need to determine future demand provokes an interest in using statistical methods based on time series to make forecasts.

At that time the supply companies had a very vertical structure, so that the producers themselves are the ones who end up marketing the energy. However, with the deregulation in the electricity markets, where the business has been divided by area of activity, into producers, distributors and marketers, in addition to a TSO, the model changed [9,10]. The need to make more accurate forecasts becomes an essential requirement. Companies multiply their efforts, and the number of proposed models increases. Figure 1 shows the evolution of the papers published in journals regarding the electricity systems. It is noticeable how an exponential growth of the publications, with almost 200 papers published in 2015 regarding this topic, from which about 50% of the publications were focused on load forecasting methods and applications.

FIGURE 1 ABOUT HERE

However, the proposed statistical models have not undergone major variations. Abu-El-Magd and Sinha [11] analyze the models used for the demand in 1982. The authors distinguish between models based on multiple regression, spectral decomposition, exponential smoothing, stochastic models, state spaces, and multivariate models. Recent studies about STLF methods [12–14] describe three main areas: fundamental, statistical and computational methods.

Only computational methods have shown a great increase in models and analyses, while "traditional" models have focused more on improving methodologies in practice [15]. The introduction of multiple seasonalities in exponential smoothing models [16,17] and in state spaces[18,19] have been a trigger to evolve these models. Some new approaches use modern methods, like Facebook's Prophet [20], covariates introduction [21] or combined with wavelets decomposition [22].

The characteristics of the short-term electricity demand series make it particularly suitable for the use of prediction models based on univariate methods [23]. The Holt-Winters methods have evolved from the original model published by Winters [24]. These models have a set of smoothing equations for the different components of the series: an equation for the level, an equation (optional) for the trend and an equation for seasonality, also optional. The prediction equation collects all the information provided by the previous equations to produce a forecast $f$ instants later. Taylor [16] proposed the double seasonal model in which the seasonal equation is divided into two nested seasonal equations, and where the prediction is also adjusted using the first-order autocorrelation error with a factor also to be determined. A generalization to nHWT is proposed [25] that is extended using discrete seasonalities, resulting nHWT-DIMS [26,27], that can be applied to irregular seasonalities [28].

A very common alternative in prediction is the use of artificial intelligence. Neural networks offer very accurate forecasts for this type of series. Among the wide range of tools available, the NARX and Shallow networks stand out. They have been the most used for this purpose. There is currently a rising trend in the presentation of models based on neural networks, as convolution neural nets (CNN) or long short-term memory (LSTM) models that take advantage of Deep Learning [29,30], with really interesting results [31]. However, the first ones are the ones used with a lot of difference.

The software developed for the prediction of electricity demand is very varied [32,33]. As a general rule, TSOs develop their own proprietary software, and rarely publish the results. The use of programs such as MATLAB® [34,35], SAS [36,37], etc. is usual for this purpose. The alternatives with free software go through the options in R, such as the MEFM (Monash Electricity Forecasting Model) package [38], and in Python the *loadforecast* library [39,40].

**3. Implemented models**

This software implements advanced forecasting methods based on nHWT, multiple seasonal decomposition and A.I. including DIMS. The following subsections describe the models implemented in the software.

*3.1. Holt-Winters*

Exponential smoothing methods were started at the 60's, with the works of Holt [41] and Brown [42]. The Holt-Winters model was developed in the 60's [43]. It consists of three smoothing equations for level $L_t$ (1), trend $T_t$ (2) and seasonality $s_t$ (3), with a forecasting equation that uses the information from the smoothing equations to make *k-ahead* forecasts (4).

$$L_t = \alpha \frac{y_t}{s_t} + (1-\alpha)(L_{t-1} + T_{t-1}) \tag{1}$$

$$T_t = \gamma(L_t - L_{t-1}) + (1-\gamma)T_{t-1} \tag{2}$$

$$s_t = \delta \frac{y_t}{L_t} + (1-\delta)s_{t-s} \tag{3}$$

$$\hat{y}_{t+k} = L_t + kT_t + s_{t-s+k} \tag{4}$$

where $s$ stands for the seasonal cycle length. It can be seen the models are recursive, so it means it is necessary to calculate initial values. Several methods are described in [44]. The smoothing parameters $\alpha$, $\gamma$ and $\delta$ are bounded in the range (0,1), and measure the weight of the smoothing, giving more importance to initial values when closer to 0, and more importance to newer when closer to 1 [45,46]. To obtain the values of the parameters, several optimization techniques can be used, always trying to minimize the one step ahead forecasting error, measured through an accuracy indicator. The most commonly used indicator used to measure the error is the root of the mean squared error (RMSE), as shown in (5).

$$RMSE = \sqrt{\frac{\sum_{i=1}^{N}(y_i-\bar{y})^2}{N}} \quad (5)$$

with N as number of observed values. The model can be expressed using additive or multiplicative trend, and with additive or multiplicative seasonality, depending on the way they are combined. Some evolutions on the model included a damping trend factor $\phi$ in trend [47,48]. Taylor introduced in the model two and three nested seasonalities [49,50], as well as the first-order autocorrelation error adjustment. These novelties outperformed previous models. García-Díaz and Trull [51] generalized the models to multiple seasonalities (nHWT) and described new initialization methods. The general model for a additive-tend and multiplicative-seasonality model is shown in (6-9).

$$L_t = \alpha\left(\frac{y_t}{\prod s_{t-s_i}^{(i)}}\right) + (1-\alpha)(L_{t-1} + \phi T_{t-1}), \quad (6)$$

$$T_t = \gamma(L_t - L_{t-1}) + (1-\gamma)\phi T_{t-1} \quad (7)$$

$$s_t^{(i)} = \delta^{(i)}\left(\frac{y_t}{L_t \prod_{j\neq i} s_{t-s_j}^{(j)}}\right) + (1-\delta^{(i)})s_{t-s_i}^{(i)} \quad (8)$$

$$\hat{y}_{t+k} = (L_t + \sum_{n=1}^{k}\phi^n T_t)\prod_i s_{t-s_i+k}^{(i)} + \varphi_{AR}^k \varepsilon_t \quad (9)$$

Here, the former seasonality $s_t$ has been split into several $s_t^{(i)}$, each with a smoothing parameter $\delta^{(i)}$. The number of seasonalities must be according to the time series under study, and $s_i$ stands for each seasonal cycle length. We included here the trend damping factor in (7) and the first autocorrelation error adjustment ($\varphi_{AR}$) in (9).

Trull et al. [52] introduced DIMS in the model, in order to deal with calendar and special events. nHWT-DIMS adds discrete seasonalities within the model that catches the seasonal effect produced by the special events and includes it as a part of the model itself, not being thus modified.

The former equations (6-9) are now modelled including DIMS as shown in equations (10-14).

$$L_t = \alpha \left( \frac{y_t}{\prod_{i=1,\ldots,n_S} s_{t-s_i}^{(i)} \prod_{h=1,\ldots,n_{DIMS}} D_{t_h^*-s_h}^{(h)}} \right) + (1-\alpha)(L_{t-1} + \phi T_{t-1}), \tag{10}$$

$$T_t = \gamma(L_t - L_{t-1}) + (1-\gamma)\phi T_{t-1} \tag{11}$$

$$s_t^{(i)} = \delta^{(i)} \left( \frac{y_t}{L_t \prod_{j \neq i} s_{t-s_j}^{(j)} \prod_{h=1,\ldots,n_{DIMS}} D_{t_h^*-s_h^*}^{(h)}} \right) + (1-\delta^{(i)}) s_{t-s_i}^{(i)} \tag{12}$$

$$D_{t_h}^{(h)} = \delta_D^{(h)} \left( \frac{y_t}{L_t \prod_{j \neq i} s_{t-s_j}^{(j)} \prod_{m=1,\ldots,n_{DIMS}} D_{t_m^*-s_m^*}^{(m)}} \right) + (1-\delta_D^{(h)}) D_{t_h-s_h^*}^{(h)} \tag{13}$$

$$\hat{y}_{t+k} = (L_t + \sum_{n=1}^{k} \phi^n T_t) \prod_i s_{t-s_i+k}^{(i)} \prod_{h=1,\ldots,n_{DIMS}} D_{t_h^*-s_h^*}^{(h)} + \varphi_{AR}^k \varepsilon_t \tag{14}$$

We introduced $D_{t_h}^{(h)}$, the new smoothing equations for the DIMS. They also include new smoothing parameters $\delta_D^{(h)}$ for each DIMS. $n_{DIMS}$ refers for the number of DIMS considered. The DIMS are integrated as a part of the model itself, thus no external modifications like dummies are needed. DIMS $h$ is only defined in times $t_h$ ($t_h \in t$). DIMS seasonal cycle length is defined by $s_h$, but the recurrence depends on the previous appearances that are not necessarily regular, must $s_h^*$ must be continuously calculated on every time step. $t_m^*$ has been defined to check whether a DIMS is defined in time step $t$ or not. Once again, depending on the way in which the seasonalities and the trend are integrated into the model, we can speak of a framework of models, as can be seen in appendix A1 of [52].

*3.2. nHWT-DIMS generalized models*

The working basis of the toolbox are the nHWT-DIMS models. One of the advantages of integrating seasonalities is the versatility with which they can do so. As a general rule, the models contemplate an integration in an additive or multiplicative way, following all the seasonalities the same generic way. However, depending on the nature of seasonality, a general model allows it to be included in one way or another.

Taking advantage of this feature of multiple seasonality, the model framework is reduced to only two generalized models, depending on the type of trend considered.

*3.2.1. nHWT-DIMS with additive trend model*

The model presented in (15-21) uses an additive damped trend, where the seasonalities are integrated so that $SA^{(i)}$ represents the additive seasonalities, $DA^{(i)}$ the additive DIMS, $SM^{(i)}$ represents the multiplicative seasonalities and $DM^{(i)}$ the multiplicative DIMS,

$$L_t = \frac{\alpha\left(y_t - \sum_i SA^{(i)}_{t-s_i} - \sum_h DA^{(h)}_{t^*_h - s^*_h}\right)}{\prod SM^{(k)}_{t-s_k} \prod DM^{(m)}_{t^*_m - s_m}} + (1-\alpha)(L_{t-1} + \phi T_{t-1}) \tag{15}$$

$$T_t = \gamma(L_t - L_{t-1}) + (1-\gamma)\phi T_{t-1} \tag{16}$$

$$SA^{(i)}_t = \frac{\delta^{(i)}\left(y_t - L_t - \sum_{j \neq i} SA^{(j)}_{t-s_j} - \sum_h DA^{(h)}_{t^*_h - s^*_h}\right)}{\prod_k SM^{(k)}_{t-s_k} \prod_m DM^{(m)}_{t^*_m - s^*_m}} + \left(1 - \delta^{(i)}\right) SA^{(i)}_{t-s_i} \tag{17}$$

$$SM^{(i)}_t = \frac{\delta^{(i)}\left(y_t - \sum_i SA^{(j)}_{t-s_j} - \sum_h DA^{(h)}_{t^*_h - s^*_h}\right)}{L_t \prod_{k \neq i} SM^{(k)}_{t-s_k} \prod_m DM^{(m)}_{t^*_m - s^*_m}} + \left(1 - \delta^{(i)}\right) SM^{(i)}_{t-s_i} \tag{18}$$

$$DA_{t_i}^{(i)} = \frac{\delta_D^{(i)}\left(y_t - L_t - \sum_j SA_{t_i^*-s_j}^{(j)} - \sum_{m \neq h} DA_{t_h^*-s_h^*}^{(h)}\right)}{\prod_k SM_{t_i^*-s_k}^{(k)} \prod_m DM_{t_m^*-s_m^*}^{(m)}} + \left(1 - \delta_D^{(i)}\right)DA_{t_i-s_i^*}^{(i)} \tag{19}$$

$$DM_{t_i}^{(i)} = \frac{\delta_D^{(i)}\left(y_t - \sum_j SA_{t_i^*-s_j}^{(j)} - \sum_h DA_{t_h^*-s_h^*}^{(h)}\right)}{L_t \prod_k SM_{t_i^*-s_k}^{(k)} \prod_{m \neq i} DM_{t_m^*-s_m^*}^{(m)}} + \left(1 - \delta_D^{(i)}\right)DM_{t_i-s_i^*}^{(i)} \tag{20}$$

$$\hat{y}_{t+k} = \left[(L_t + k\,\phi T_t) + \left(\sum_i SA_{t-s_i+k}^{(i)} + \sum_h DA_{t_h^*-s_h^*+k}^{(h)}\right)\right] \prod_n SM_{t-s_n+k}^{(n)} \prod_m DM_{t_m^*-s_m^*+k}^{(m)} + \varphi_{AR}^k \varepsilon_t \tag{21}$$

This formulation allows the combination of all kinds of seasonality. If it is not desired to consider the damping of the trend, the effect of the parameter $\phi$ can be removed by assigning it a value of 0. The same will happen with the adjustment of the first-order autocorrelation error ($\varphi_{AR} = 0$).

*3.2.2. nHWT-DIMS with multiplicative trend model*

In case the way of integrating the trend is multiplicative, the previous formulation is not correct. It is necessary to build a new model. It would be possible to generalize in a single model all the equations, but it would result too complicated to work with. Therefore, it has been split into two models. The new model for multiplicative trend is shown in equations (22-28), where $R_t$ is the multiplicative trend.

$$L_t = \frac{\alpha\left(y_t - \sum_i SA_{t-s_j}^{(j)} - \sum_h DA_{t_h^*-s_h^*}^{(h)}\right)}{\prod_k SM_{t-s_k}^{(k)} \prod_m DM_{t_m^*-s_m^*}^{(m)}} + (1-\alpha)\left(L_{t-1} R_{t-1}^\phi\right) \tag{22}$$

$$R_t = \gamma(L_t/L_{t-1}) + (1-\gamma)R_{t-1}^\phi \tag{23}$$

$$SA_t^{(i)} = \frac{\delta^{(i)}\left(y_t - L_t - \sum_{j \neq i} SA_{t-s_j}^{(j)} - \sum_h DA_{t_h^*-s_h^*}^{(h)}\right)}{\prod_k SM_{t-s_k}^{(k)} \prod_m DM_{t_m^*-s_m^*}^{(m)}} + \left(1 - \delta^{(i)}\right)SA_{t-s_i}^{(i)} \tag{24}$$

$$SM_t^{(i)} = \frac{\delta^{(i)}\left(y_t - \sum_j SA_{t-s_j}^{(j)} - \sum_h DA_{t_h^*-s_h^*}^{(h)}\right)}{L_t \prod_{k \neq i} SM_{t-s_k}^{(k)} \prod_m DM_{t_m^*-s_m^*}^{(m)}} + (1-\delta^{(i)}) SM_{t-s_i}^{(i)} \tag{25}$$

$$DA_{t_i^*}^{(i)} = \frac{\delta_D^{(h)}\left(y_t - L_t - \sum_j SA_{t-s_j}^{(j)} - \sum_{h \neq i} DA_{t_h^*-s_h^*}^{(h)}\right)}{\prod_k SM_{t-s_k}^{(k)} \prod_m DM_{t_m^*-s_m^*}^{(m)}} + (1-\delta_D^{(i)}) DA_{t_i^*-s_i^*}^{(i)} \tag{26}$$

$$DM_{t_i^*}^{(i)} = \frac{\delta_D^{(h)}\left(y_t - \sum_j SA_{t-s_j}^{(j)} - \sum_h DA_{t_h^*-s_h^*}^{(h)}\right)}{L_t \prod_k SM_{t-s_k}^{(k)} \prod_{m \neq i} DM_{t_m^*-s_m^*}^{(m)}} + (1-\delta_D^{(i)}) DM_{t_i^*-s_i^*}^{(i)} \tag{27}$$

$$\hat{y}_{t+k} = \left[L_t R_t^{\sum_{l=1}^k \phi^k} + \left(\sum_i SA_{t-s_i+k}^{(i)} + \sum_h DA_{t_h^*-s_h^*+k}^{(h)}\right)\right] \prod_n SM_{t-s_n+k}^{(n)} \prod_m DM_{t_m^*-s_m^*+k}^{(m)} + \varphi_{AR}^k \varepsilon_t \tag{28}$$

*3.3. Additional implementations: Neural networks*

The library also includes neural network models. It has been considered interesting to include a prediction module using neural networks, in which the use of DIMS is included as a novelty.

The artificial intelligence models are widely used in the prediction of electricity demand due to their good forecasts [31,53]. The toolbox includes NARX and shallow neural network models. A schematic of the neural network used is shown in the Figure 2. There is an input layer to which the observed values and variables are input. The network is organized into nodes and connections, similar to the human brain, which is why they are known as neurons and synapses. The nodes are organized in layers performing simple operations in parallel, using the information of neurons from previous layers through connections. Input variables are connected to an input layer. The following layers are known as hidden layers, and in them the number of neurons and the weight of the connections are adjusted. The last layer, the output layer, is the one that produces the prediction of the network.

FIGURE 2 ABOUT HERE

The input variables $x_t$ are exogenous variables that can be used to improve the model. The variables are connected by means of axioms with weights $w_i$ assigned during the training, and with an activation function $f(.)$ that integrates with an aggregation function $\Sigma$. $b$ represents the bias. The output $\hat{y}_{t+1}$ provides forecasts after the network has been trained.

The mathematical representation of the neural network model can be seen in equation (1). The values of $D_x$ and $D_y$ represent the delays that are applied to the variables in the network.

$$\hat{y}_{t+1} = f\left[x_t, x_{t-1}, \dots, x_{t-D_x}, y_t, y_{t-1}, \dots, y_{t-D_y+1}\right] \qquad (1)$$

Obtaining the weights of the networks is obtained through a training process. The Levenberg-Marquardt algorithm is used in the toolbox, although it is possible to use others. Error minimization is done through the mean square error (MSE).

The preparation of the data for the input layer is necessary. According to [34], the time, day of the week, and weather information (if available) are included as additional information. The series used present several seasonalities, so it is necessary to include a recurrence to previous values through the use of moving averages, one for each different seasonality, and of amplitude the length of the considered seasonal cycle. To include special events due to the calendar effect, a dichotomous variable is generally used that indicates whether it is a holiday or a working day. The use of DIMS allows not only to indicate the presence of a holiday, but also to consider its effect on the series. DIMS are included in the model as a new exogenous variable, in a discontinuous way, which includes a seasonal pattern only at the times when the special event occurs. Therefore, as many DIMS series as discrete seasonalities have been incorporated into the model are included in the input layer.

**4. Architecture of the toolbox**

Due to the large data size of the series used, it was decided that the best platform to use to implement the models was MATLAB®. The programming uses a proprietary software language, although it is similar to others such as C++, Java, etc. However, its use provides a series of advantages: there are already developed and verified libraries, and memory management is exceptional. The library has been developed using an object-oriented architecture (OO) based on the *mts* class (included as a package), which is a subclass based on MATLAB *timeseries* class. The use of *timeseries* allows reuse of a large number of functions offered by MATLAB.

From this class, inherited classes have been generated with specific purposes: to manage the database from which to extract the data, and to carry out the planned analyses. Additionally, additional classes have been designed to perform specific actions.

*4.1. Class diagram of the toolbox*

The structure of inheritance and derivation of classes can be seen in the class diagram shown in the Figure 3. In this diagram, the object of each class has been organized by color. Orange shows the timeseries class, typical of MATLAB, from which *mts*, the base class of the library, inherits. The classes whose purpose is to manage the databases are presented on a green background, and the classes destined to perform time series analysis are shown in blue. Class dependencies are indicated by arrows.

*FIGURE 3 ABOUT HERE*

*4.2. Class mts*

The mts class extends the functionality of timeseries to include multiple seasonalities. This class is in charge of managing the seasonalities of the model, in addition to the discrete seasonalities. mts works as a container for the covariates of the model.

The methods included in this class are organized as follows: on the one hand, there are the methods aimed at managing the data, on the other, the methods for performing prediction error calculations, and finally, the methods for performing operations with the series. Table 1 shows the main properties and methods related to this series.

The *mts* class is a container in addition to the DIMS. These elements have a particular structure in order to contain the necessary information. They are first identified, and the seasonal length is determined. Then, it is indicated in which moments of the series they are defined, indicating the first moment in each of the interventions.

TABLE 1 about here

Classes derived from *mts* overload the Seasons and the DIMS containers to give them additional features, depending on the goal. Thus, the *mshw* class will add to the DIMS the properties of how seasonality will be related in the model (additive or multiplicative) and how its initial value will be obtained. TSCov contains any exogenous variables that may be needed for the models.

Among the prominent methods in the mts class are the functions *stl*, *mstl*, *stlplot*. These functions perform a decomposition based on Loess [54,55] of the series into trend and seasonal (multiple) components. The decomposition uses an algorithm so that all the seasonalities are obtained simultaneously and prior to obtaining the DIMS decomposition. The discrete seasonalities are obtained using the methodology explained in [56].

Figure 4 shows an example of multiple seasonal STL decomposition plot by the *stlplot* function. Figure 4a shows the regular multiple seasonal decomposition, where the original data and the regular seasonalities are decomposed. The remainder is also shown below. Figure 4b has a different layout. The seasonal decomposition for the DIMS is shown at the left side of the figure, while the appearances of the DIMS in the time series is shown at the right side. The code needed to generate the graphs is included in Appendix A.

FIGURE 4 ABOUT HERE

From the STL decomposition predictions can be made keeping the structure of the decomposition. It is a very brief way of predicting, and it is of no more interest than the application's own internals. However, it is indicated because it may be useful to the reader.

*4.3. Class Elecmts*

The data to be analyzed can be entered from any data source, including a numeric array. However, much of the data used in the library and applied in previous papers is provided through the different data wrapper classes. The *Elecmts* class handles the loading of data provided by TSOs from different countries.

Since this is a class derived from mts, you can access the existing properties and methods of that class. This class provides the following methods to be able to work with the included databases. The *ListDB* function allows the user to know the databases included in the data file. The *ListSeries* method allows you to see the series included in each database.

Table 2 shows the main databases available in the toolbox. Databases are identified by a word, as indicated in the first column. The second column is the official name given to the database, while the third column describes its content. Subsequently, each database has its own structure, according to the available data. Because the size of the files is considerable, each database manages its own file.

To load the data, the user must create an object of type *Elecmts* indicating the database with the *db* parameter, and with the different parameters necessary to locate the series to be used. You can also indicate the period or the time window to load. If the loading of several data series is required, the library will load the data as covariates.

TABLE 2 ABOUT HERE

*4.4. Class Climmts*

The *Climmts* class is a wrapper and database manager for the mts class related to weather variables. Like *Elecmts*, it is responsible for managing the data load in the *Climmts* class object. The *ListDB* and *ListSeries* methods work the same way as in the previous case. Table 3 lists the current database available.

TABLE 3 ABOUT HERE

*4.5. Class Demmts*

Demographic data can be interesting to calculate models with better precision. This class derived from *mts* manages the library's demographic database. At present, demographic data is only available for Spain and some for the USA. The functions used for its management are the same as in the previous cases.

This class also includes tourism indicators used in previous works, such as the Human Pressure Indicator –both daily and monthly–, provided by the Balearic Institute of Statistics (IBESTAT).

*4.6. Class mshw*

The *mshw* class provides a series of tools to forecast time series using multiple seasonal Holt-Winters models, including the possibility of including DIMS within the model. This class derives directly from *mts*, but some of the inherited processes are modified to suit the needs of the models. The Seasons property, for example, includes in addition to the length of the cycle of each seasonality, the method used to calculate the effect of the seasonality, being additive or multiplicative. The way to calculate the initial values of each seasonality is also included.

The smoothing parameters (alpha, gamma, delta, phi and ar1) are managed from this class. Among the methods to highlight in this class, the *FindParams* method stands out. This method obtains the optimized values of the smoothing parameters by minimizing the prediction error. The algorithm used in the solve de minimization problem, as well as the error indicator can be customized. Although simplex method is commonly used, genetic algorithms, pattern search and others are available to be used in order to find the parameters.

Table 4 shows the most important properties and methods of this class. To make predictions, the forecast function offers a wide range of possibilities. If the parameters have been optimized, the call to this function will indicate the number of predictions to make.

The *mshw* class implements a wide variety of Holt-Winters models, as described in Section 3. It also introduces different methodologies for obtaining seeds, both for the initial values of the models and for the initial values of the DIMS. The *mshw* object is created (usually inheriting from *Elecmts* object). The model structure is defined by setting the features of the trend and the seasonalities. The user can then enter the smoothing parameters' values or leave the toolbox to obtain them. The forecasts can be then produced and plotted.

We show an example of the forecasting properties using *mshw* in Figure 5. The instructions list used to obtain the forecasts is listed in Appendix B.

TABLE 4 ABOUT HERE

FIGURE 5 ABOUT HERE

*4.7. Class msnn*

The implementation of A.I. have also been considered in this toolbox. The *msnn* class manages specialized neural network models for prediction. It is a class derived from *mts*, and has the same properties and methods. Additionally, it has a neural network container that must be trained using the *FindParams* method.

The neural network used can include covariates, which must be correctly prepared. The *forecastoption* and *Covforecastoptions* properties allow you to establish the working conditions of the neural network.

*4.8. Helper classes*

The *Calmts* class is a helper class that manages the database of events in the calendar, and provides the information needed to generate DIMS. The *optim* class is designed to contain the information needed during the parameter fetching process.

*4.9. Graphical User Interface*

In order to facilitate the user's work with the toolbox, a graphical user interface (GUI) has been developed. Using this GUI you can perform the basic functions of work with the objects described, since it allows you to obtain data from database sources, display the series and perform the relevant analysis.

FIGURE 6 ABOUT HERE

## 5. Application

To check the effectiveness of the toolbox presented here, application cases already published in previous works will be shown. In each case, the objective of the application is explained, and the effectiveness of the result of the forecasts will be verified by measuring the error in the forecasts. To do this, the Mean Average Percentage Percentage Error (MAPE%) will be used.

*5.1. Spanish Electricity Demand Forecasting during Easter*

The prediction of electricity demand in holiday periods usually carries with it the problem that the models are not specially designed for this situation. This causes the effectiveness of the forecasts to decline sharply. In order to deal with this situation, the models use modifiers such as dummy variables, which consider each particular case as a unique model. On other occasions, this special situation condition is simply analyzed and eliminated, and after applying the known prediction methods, the modification made is incorporated again.

The use of DIMS allows the singularity of special events to be introduced into the model itself, and where appropriate, for Easter, a festive period that lasts several days and with high variability. In Spain, Holy Friday is a holiday; Holy Thursday is a holiday depending on the region and the Monday after Easter in some regions it is a holiday, and in others, depending on the year, it is. This generates a particular situation of variability that the usual models fail to work correctly. DIMS, however, adapt naturally by being introduced as one more seasonality, albeit discreetly. In [19] a comparison was made using a double seasonal model, a TBATS (Exponential smoothing state space model with Box-Cox transformation, ARMA errors, Trend and Seasonal components) without modifying the series, and another modifying the series as shown. indicates in, a Shallow neural network model, and an nHWT-DIMS model.

The indicator used to compare the results obtained is the Mean Absolute Percentage Error (MAPE%), which is the most commonly used for this purpose. Figure 7shows the results of the forecasts of the best models, where it can be seen how the nHWT-DIMS and the ANN stand out from the rest of the models.

FIGURE 7 ABOUT HERE

*5.2. STLF in holidays and bridges*

Following the same line of work, the use of DIMS was extended to the entire range of holidays and bridges that exist in the Spanish and French calendars [27]. Unlike the previous case, where there is a discrete seasonality that appears only once a year, we find on this occasion how a special event (holidays) occurs up to 9 or 10 times in a year, and another event (bridges) that is enormously volatile and occurs 3 or 4 times a year.

The variability that this situation provides means that the usual models resort to modeling these events as singular in which an additional treatment of the series, or an expressly rectified model make good predictions. The use of DIMS allows these anomalous demands to be integrated as part of the model itself. Distinguishing the type of event a priori, the model adjusts to the series without the need for modifications or filters, and makes predictions based on real information. Figure 8 shows a comparison of the results obtained using nHWT-DIMS and other methods described in [27]. The figure shows the MAPE(%) of 24-hours-ahead forecasts during special days. The subfigure (a) shows the results for the Spanish hourly electricity demand in the holidays (identified by name) and bridges (written as date). The models $AMC_{24,168,Easter,Holidays,Bridges}$ and $NMC_{24,168,Easter,Holidays,Bridges}$ show a better performance against the rest, with MAPEs under 5%. Only the best methods have been reported here.

The subfigure (b) shows the results of 48-midhours-ahead forecasting for special days in France, identified by their name. It has been compared the results obtained to those reported by [57].

FIGURE 8 ABOUT HERE

**6. Conclusions and future research**

This article presents a new toolbox based on MATLAB® whose objective is the prediction of electrical demand using exponential smoothing methods with multiple seasonal Holt-Winters models. Neural networks are also available for benchmark purposes.

Firstly, the general formulation for the nHWT models has been presented, both with additive and multiplicative trend, including damping. The model is presented valid so that the different seasonalities are included in the model in an additive or multiplicative way, independently of the rest. In the same way, the use of DIMS within the model allows situations with special events to be modeled internally, without the need for external modifications with exogenous variables.

Secondly, the library itself has been described. On the one hand, the architecture of the library and its organization in classes are shown. The methods and properties associated with the different classes are described in depth, as well as their relationship with the parameters of the models. The article hovers over the most important, although sufficient to be able to carry out accurate predictions.

To demonstrate the functionality of the toolbox, tested and published application cases are included, in which the excellent results offered by the toolbox demonstrate the interest of its use. On the one hand, the application of the DIMS for forecasting demand in the Easter period has been shown, and then applied to all holidays (including long weekends).

This toolbox has been mainly used for short-term electricity demand prediction. Its formulation allows its use for longer periods, although it has not been optimized for this. The data used has been provided by the TSOs through their websites.

Future work related to this toolbox is the automation of selection processes for special events and the inclusion of new models. The use of DIMS has been shown to be effective, so they will be applied to new prediction methods.

APPENDIX A

The above notation can be complex, therefore it has been decided to use a section to explain the terms. First, Table 5 shows the equations of the model. The first column indicates the notation of the equations, while the rest of the columns describe it. All equations are in function of time $t$. It is important to note that the seasonal indices have an associated superscript that indicates the ID of the seasonality. For example, if we use intraday seasonalities, we could use $(j) = (24)$ due to the seasonal length, or just $(j) = diary$. This is the case of seasonalities that are included in the model in an additive way, but for multiplicative ones it would be denoted in the same way. The same goes for DIMS, although these are generally associated with events, such as Easter, Holidays, etc.

TABLE 5 ABOUT HERE

The time subscript that accompanies the DIMS is different and includes a reference to the DIMS itself. The notation $t_i^*$ or its equivalents indicate that the time variable $t$ is used, but only if the DIMS is defined for that instant. Similarly, the subscripts do indicate the length of the seasonal period considered, being, for example, for intraday seasonality $s_{24} = 24$. In the case of DIMS, it works differently. The subscript is relative to the last occurrence of the DIMS prior to this instance, and therefore the length is determined at each occurrence. Trull et al. [27] describe this process. The equation $\hat{y}_t(f)$ provides the $f$-ahead point forecasts starting from the las observed value.

The equations also include smoothing parameters, which are described in the Table 7. For each seasonality considered, whether regular or DIMS, a delta described with superscripts will be used in the same way as in the previous case. The value of these parameters is bounded between 0 and 1, since in this way they respond stably to the models.

Since the models are recursive, the initial values of the series must be obtained previously. The use of multiple seasonalities and, even more so, the use of DIMS, requires a complex process of obtaining parameters. The toolbox allows to use different methods according to [58], which the user decides.

Finally, the influence of the value of the parameters on the predictions is fundamental. The process of optimizing the parameters is done through algorithms to minimize the prediction error in the set used for the adjustment. The toolbox allows the use of algorithms based on the Nelder-Mead simplex[59], genetic algorithms, BFGS and L-BFGS-B, among others [60].

TABLE 7 ABOUT HERE

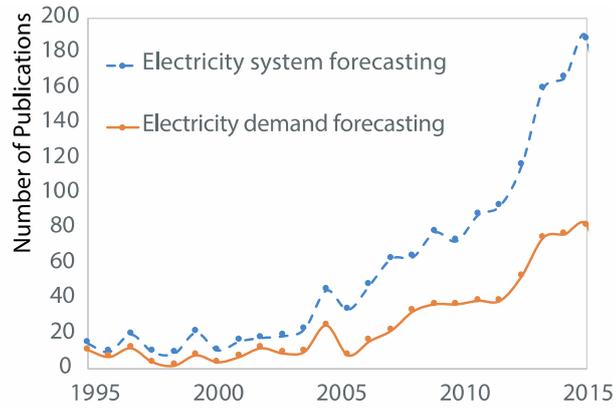

Figure 1. Evolution of the number of publications related to electricity systems and focused on electricity demand forecasting. Source: own survey in JCR-indexed journals.

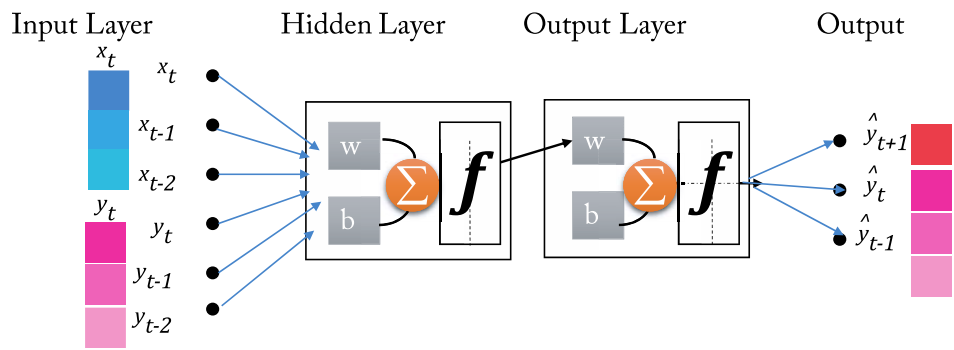

Figure 2. Neural network diagram.

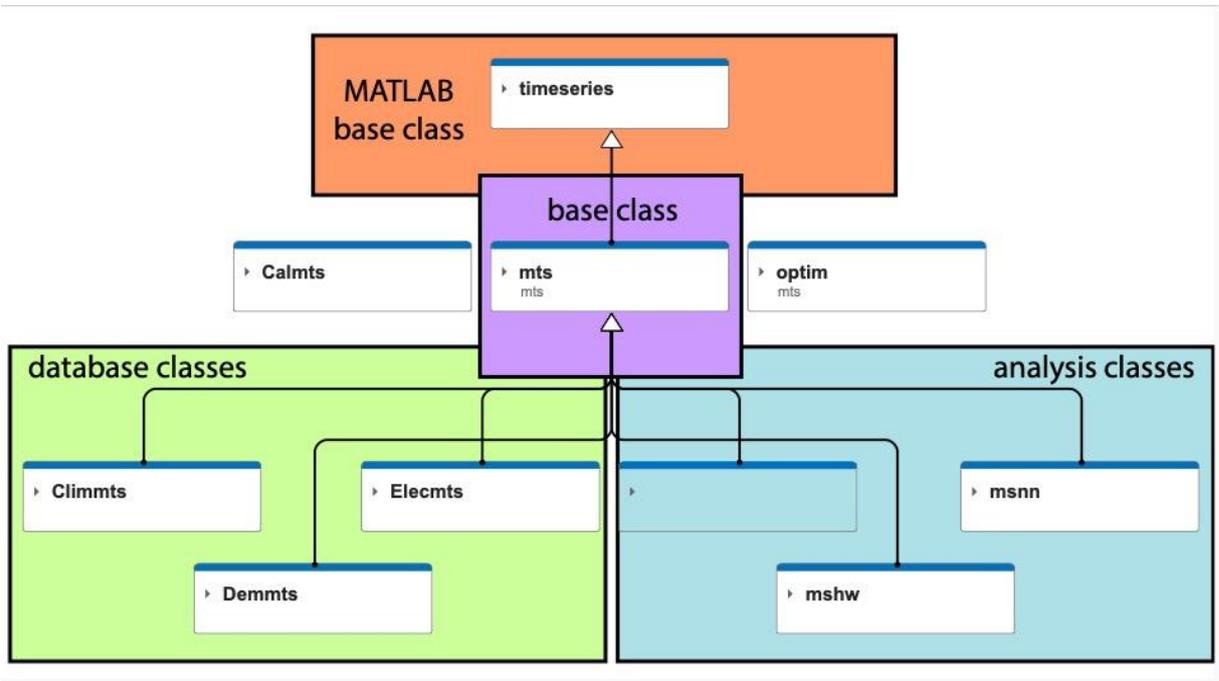

Figure 3. Class diagram of the toolbox

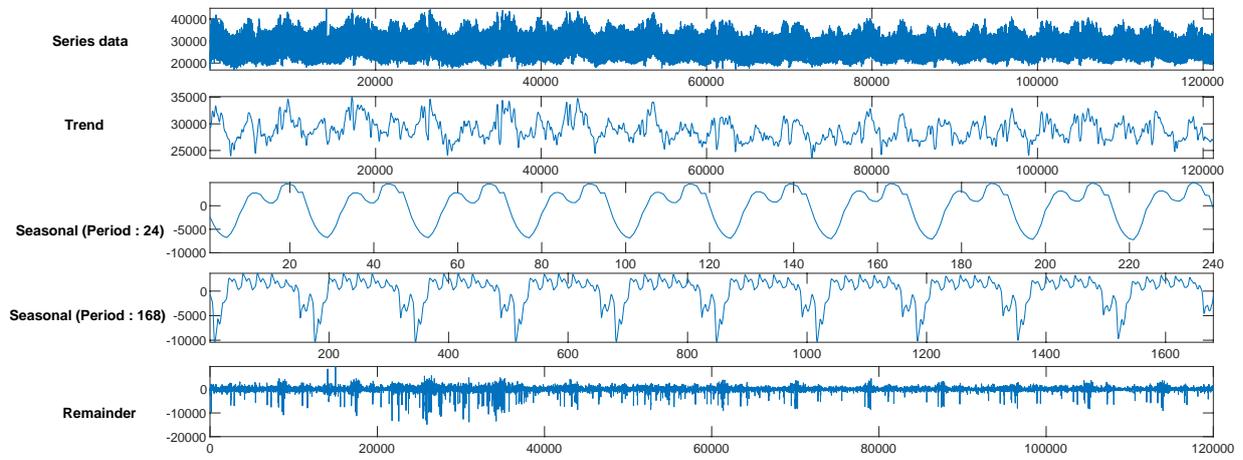

(a)

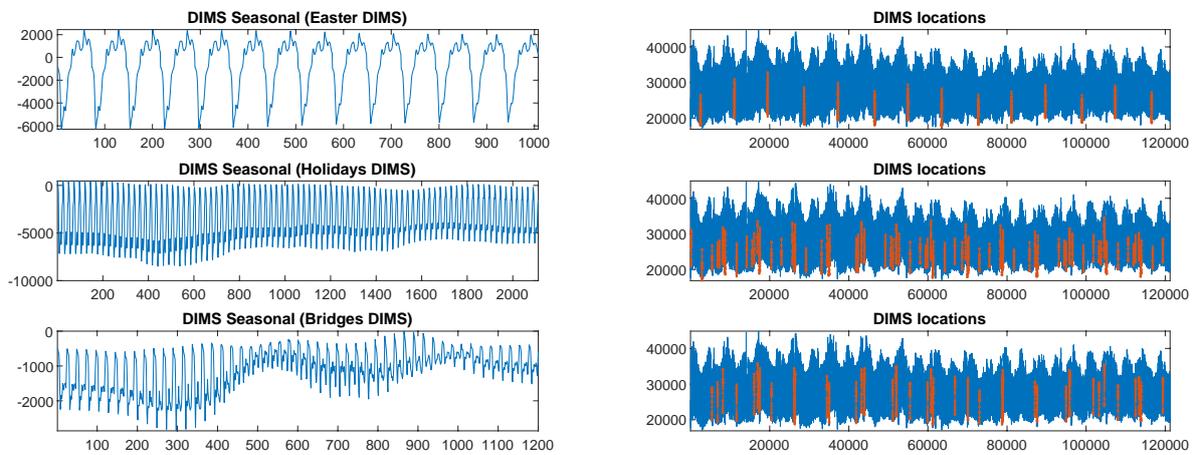

(b)

Figure 4. MSTL plot produced by the toolbox. (a) Regular seasonal decomposition, and remainder. (b) DIMS seasonal decomposition with the DIMS location in the time series.

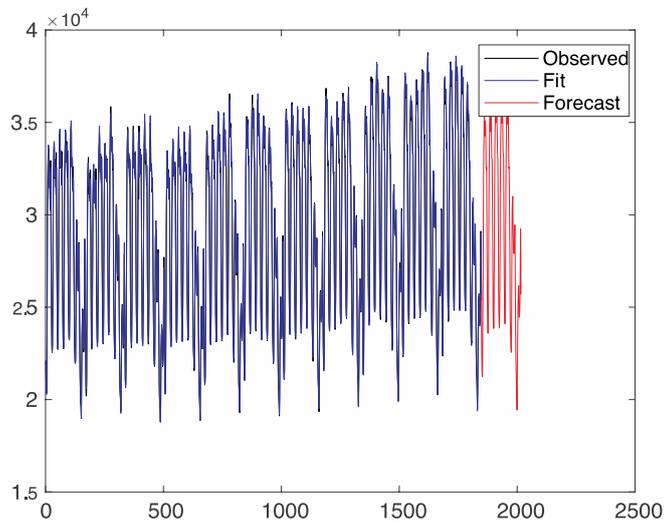

*Figure 5. forecasts produced by mshw object after optimizing the parameters.*

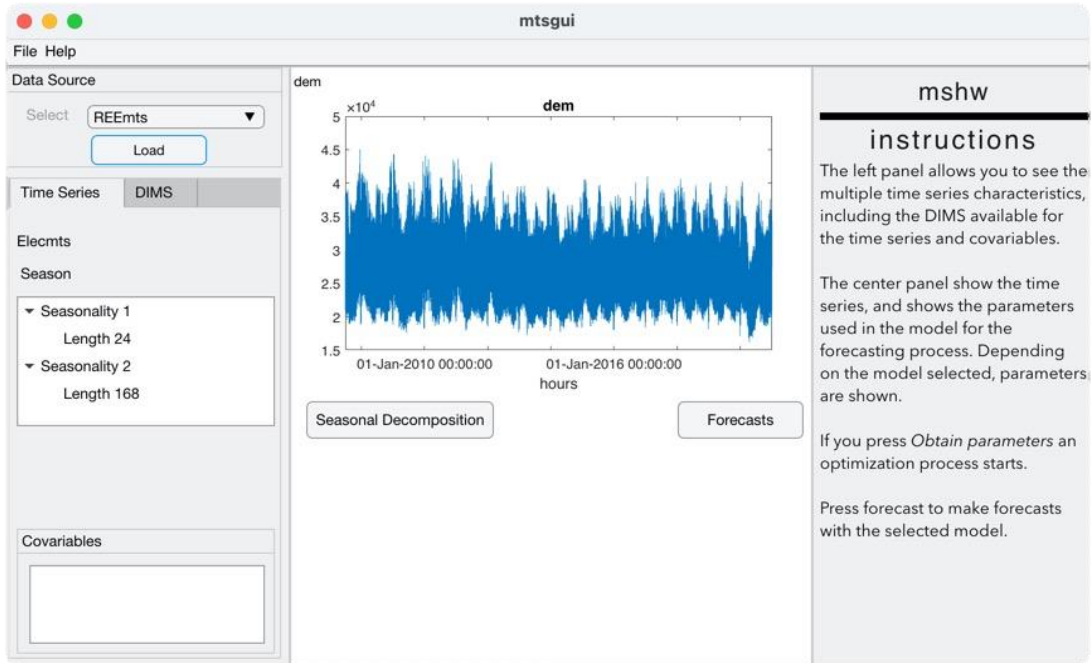

*Figure 6. mtsgui is a graphical user interface to easily manage mts objects.*

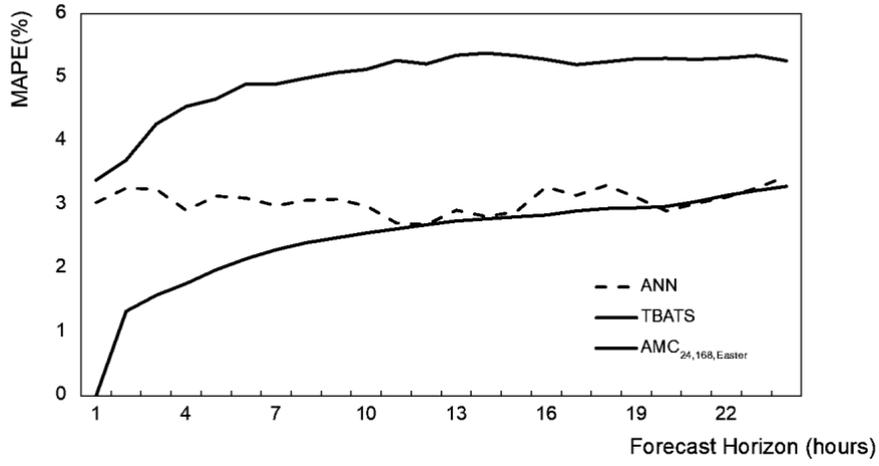

Figure 7. Comparison of the accuracy while forecasting from several methods.

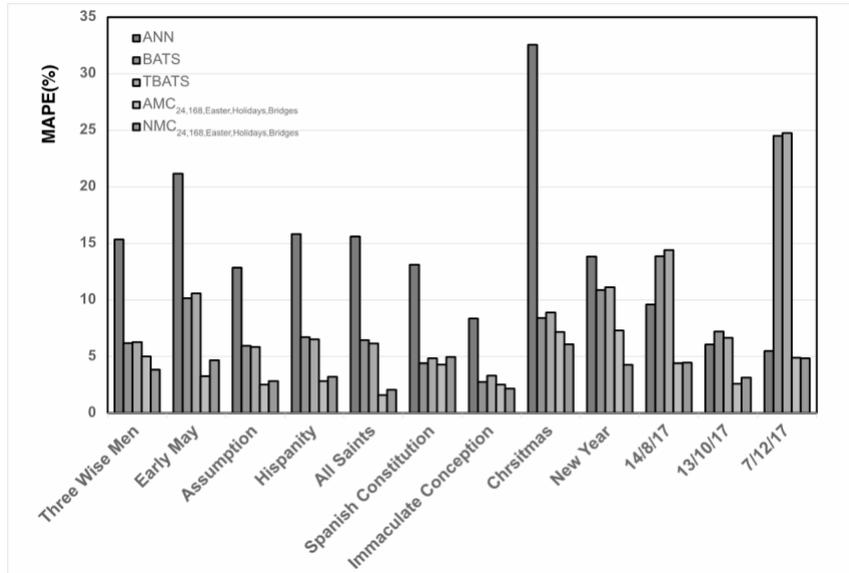

(a)

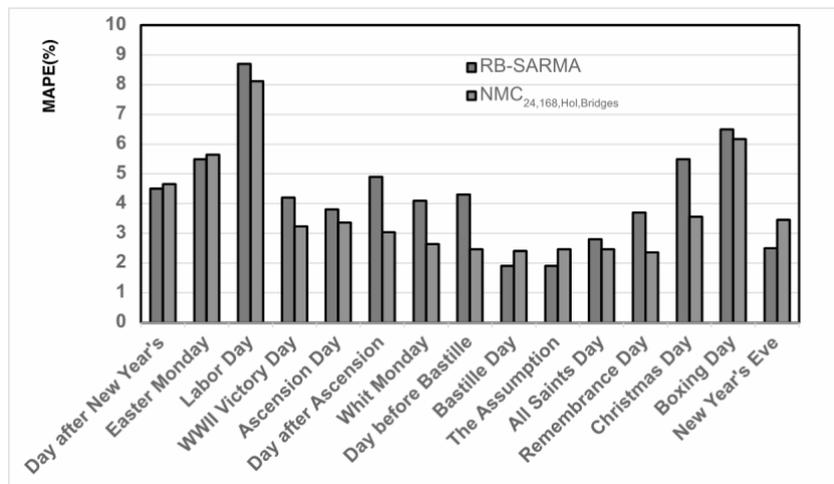

(b)

Figure 8. Accuracy of the forecasts produced for (a) Hourly Electricity Demand and (b) Half-hourly electricity demand. Measured in MAPE.

Table 1. Properties and methods for the class *mts*.

| Property | Description | Data Type | Method | Description |
|---|---|---|---|---|
| DIMS | DIMS description array | Struct | addDIMS, getDIMS, modifyDIMS, removeDIMS | DIMS management in the mts object |
| Seasons | Seasons definition | Array / Struct | addCov, removeCov | Covariate management in the mts object |
| TSCov | Covariates | TSCollection | aic, ape, rmse, mape, etc… | Performance indicator calculation, given two mts object |
| | | | mstl, stl, stlplot | Seasonal decomposition using Loess, including multiple seasonal and plotting |

Table 2. Main electricity databases available in the library.

| Identifier | Name | Description |
|---|---|---|
| REE | Red Eléctrica de España | Data from Red eléctrica de España |
| ENTSO | European Network of Transmission System Operators" | ENTSO-E, the European Network of Transmission System Operators, represents 43 electricity transmission system operators (TSOs) from 36 countries across Europe, thus extending beyond EU borders. |
| DNO | Distribution Network Operators in the UK | Data from the Distribution Network Operators in the UK |
| FEPC | Federation of Electric Power Companies (FEPC) | Data from the Federation of Electric Power Companies (FEPC), Japan |
| RTE | RTE, the French transmission system operator | Data from RTE France |
| US_ISO_NY | The NYISO is the New York Independent System Operator | Data from NY ISO |
| TERNA | Italian Transmission System Operator | DATA from TERNA Italy |
| AEMO | Australian Energy Market Operator | Data from AEMO (NEM data) |

Table 3. Climatological databases included in this software.

| Identifier | Name | Description |
|---|---|---|
| AEMET | AEMET OpenData | AEMET's Spanish Climate Data |
| wunderground | Weather Underground Data | The Weather Company, LLC, an IBM Business ("WUI") |

Table 4. Equation definitions.

| Equation | Description | Comments |
|---|---|---|
| $y_t$ | Observed values | |
| $S_t$ | Level | |
| $T_t$ | Trend | Additive trend |
| $R_t$ | Trend | Multiplicative trend |
| $IA_t^{(j)}$ | Seasonal Indices | Additive seasonal indices. *j* indicates de id of the seasonlity. |
| $IM_t^{(k)}$ | Seasonal Indices | Multiplicative seasonal indices. *k* indicates de id of the seasonlity |
| $DA_{t_h^*}^{(h)}$ | DIMS Indices | Discrete Additive seasonal indices (DIMS). *h* indicates de id of the discrete seasonlity |
| $DM_{t_h^*}^{(m)}$ | DIMS Indices | Additive seasonal indices. *m* indicates de id of the seasonlity |
| $\hat{y}_t(f)$ | Forecasting | Forecasting equation. *f*-ahead forecasts. |

Table 5. Properties and methods for the class *mshw*.

| Property | Description | Data Type | Method | Description |
|---|---|---|---|---|
| alpha, gamma | Smoothing parameters for level and trend | [0,1] | InitValues | Obtains the seeds of the model. |
| deltas | Seasonal smoothing parameters | [0,1] | FindParams | Obtains optimized smoothing parameters for the model, and initial values through InitValues. |
| deltaD | DIMS smoothing parameters | [0,1] | forecast | Performs a forecast |
| phi | Trend damping factor | [0,1] | mforecast | This method is used to provide multiple forecasts, organized in a grid of forecasts. |
| ar1 | ar(1) adjustment factor | | accuracy | Shows the accuracy obtained while fitting the model and obtaining the parameters |
| Seasons | struct | | | |

Table 6. Parameters used in the equations.

| Parameter | Description | Comments |
|---|---|---|
| $\alpha$ | Level smoothing parameter | |
| $\gamma$ | Trend smoothing parameter | |
| $\delta^{(i)}$ | Seasonal indices smoothing parameter | $i$ indicates the id of the seasonality |
| $\delta_D^{(h)}$ | DIMS smoothing parameter | $h$ indicates the id of the discrete seasonality |
| $\varphi_{AR}$ | AR(1) error adjustment factor | |
| $\phi$ | Trend damping factor | |